# Fuzzy Inference Systems Optimization

Pretesh Patel[1] and Tshilidzi Marwala[1]
[1]*University of Johannesburg
Faculty of Engineering and the Built Environment
South Africa*

**Abstract**
This paper compares various optimization methods for fuzzy inference system optimization. The optimization methods compared are genetic algorithm, particle swarm optimization and simulated annealing. When these techniques were implemented it was observed that the performance of each technique within the fuzzy inference system classification was context dependent.

## 1. Introduction

Satisfied customers establishes loyalty, provides opportunities of selling additional products and services. Satisfied customers also reduce the probability of losing business to competitors. However, customer dissatisfaction results in direct revenue losses due to customer churn as well as damage to business reputation. Therefore, the improvement of customer experience is a vital priority for contact centres across all industries.
Interactive Voice Response (IVR) systems are used by businesses to provide customers with a convenient, consistent and reliable contact channel to access information fast. When these systems are implemented correctly, they assist in improving customer experience (Sharma & Kunins, 2002). An IVR system, an automated telephony system, interacts with callers to gather relevant information and to route calls to the appropriate destinations (Sharma & Kunins, 2002). This system inputs comprise of voice, Dual Tone Multi-Frequency (DTMF) keypad selection or a combination of voice and DTMF. IVR system outputs are appropriate responses in the form of voice, fax, callback, e-mails and other media (Sharma & Kunins, 2002). These systems may consist of telephony equipment, software applications, databases and supporting infrastructure. A major objective of an IVR system is to improve customer experience, while reducing operating costs. This is achieved by automating transactions that are generally conducted with the aid of a Customer Service Agent (CSA).
Today, customers are exposed to many businesses that provide excellent services. The interactions with these businesses set customer expectations. As a result, in order to remain competitive in the current market, all customer-facing technology should be analyzed to ensure that these systems support business service strategies and deliver the preferred customer experience.

In order to provide these essential capabilities, contact centres employ business analytics for IVR solutions to scrutinize their systems. These solutions compute performance measures that determine the number of tasks completed successfully, the number of callers that selected to transfer to CSA and the number of caller disconnects. Additionally, further details such as the tasks where caller disconnects or transfers to CSA occurred are also computed. The transfer to CSA reason such as system errors or caller preferring to interact with CSA is also provided. Currently, business analytics for IVR solutions that compute implementation detail measurements such as speech recognition accuracy per task as well as out of grammar or invalid option selections can also be found (Miller, 2007).

The aim of this research is to develop a call classification system, using computational intelligent techniques, which could aid businesses in quantifying caller behaviour within their automated IVR applications. It is anticipated that this system would be used in conjunction with other customer behaviour analysis techniques such as examining call recordings. As a result, this system should be used to confirm the IVR system performance in relation to customer interaction.

Voice Extensible Markup Language (VXML) is used to create IVR applications. These applications comprise of voice-based dialog scripts that consist of form or dialog elements. The form or dialog elements are used to group input and output sections together. A field element is used to obtain, interpret and react to caller input information. Further information on field elements can found in (VoiceGenie, 2011). In this research, a task such as 'Say Account' corresponds to a field element, 'Say Account'. The form and dialog elements contain field elements (VoiceGenie, 2011).

Similar research has been conducted (Patel & Marwala, 2010). The research determined that the Artificial Neural Network (ANN) and the Support Vector Machine (SVM) techniques outperformed the Fuzzy Inference System (FIS) technique in caller behaviour classification. In order to improve the performance of the FIS technique in this classification application, this research examines the use of Genetic Algorithm (GA), Particle Swarm Optimization (PSO) and Simulated Annealing (SA) algorithms to optimize the membership functions and set of fuzzy inference rules employed within the model.

GA and PSO are evolutionary heuristics that are population-based search methods. These evolutionary methods conduct an exploration of the search space to identify optimum solutions by employing a form of direct random search process. Due to these techniques considering a population of candidate solutions rather than a single candidate solution, GA and PSO differs from conventional optimization techniques. PSO has been motivated by the swarming or collaborative behaviour of biological populations. GA has been inspired by the mechanism of natural evolution. These are effective global search methods that have been applied to various problems across different industries (Jones, 2006; Panduro et. al, 2009).

In contrast, the SA is an iterative approach that continuously updates a single candidate solution until a termination condition is achieved. SA methods belong to the probabilistic hill-climbing class of algorithms that dynamically alter the probability of accepting inferior cost or fitness values (Romeo & Sangiovanni-Vincentelli, 1985). SA has also been used in a range of problems across different industries (Ethni et. al, 2009; Manikas & Cain, 1996).

The sections to follow examine the classification system and its implementation methodology. An explanation of the FIS classifier system follows. Thereafter, the implementations of the GA, PSO and SA optimization techniques are described. The paper ends with the comparison of the results of the optimization methods considered and the selection of the appropriate technique.

## 2. Caller behaviour classification system

The objective of the classification system is to provide caller behaviour metrics that can assist contact centres in identifying areas of improvement within IVR applications. The developed system comprises of field classifiers that are trained on data extracted from IVR log event files. These files are generated by the IVR platform and comprise of specific event data that is produced during a call to the automated system. Event data such as call begin, form enter, form select, automatic speech recognition data, transfer to CSA data and call disconnect data are written to the log files (VoiceGenie, 2007).

Table 1 illustrates the inputs and outputs of the field classifiers. These inputs are utilized to characterize the caller experience at a field within an IVR application. The outputs of the classifier are interaction classes that summarize the caller behaviour.

| Inputs | Outputs | Output Interaction class |
|---|---|---|
| Confidence | Field performance | Good, acceptable, investigate, bad |
| No matches | Field transfer reason | Difficulty, no transfer, unknown |
| No inputs | Field caller disconnect reason | Difficulty, no transfer, unknown |
| Max speech timeouts | Field difficulty attempt | Attempt 1, attempt 2, attempt 3 |
| Barge-ins | Field duration | High, medium, low |
| Caller disconnect | Field recognition level | High, medium, low |
| Transfer to Customer Service Agent | Experienced caller | True, false |
| DTMF transfer | | |
| Duration | | |
| System error | | |
| Confirmed | | |

Table 1. Inputs and outputs of the field classifier

The confidence input, a percentage value, represents the IVR speech recognition probability. The larger the confidence value, the greater the probability the system interpreted the caller successfully. IVR applications are developed to accommodate a finite number of caller responses to automated questions such as "How much would you like to pay?" If the caller response to the question is not within the anticipated answers, a no match event occurs. These events are represented by the no match inputs. In this research, the IVR application considered provides a maximum of 3 no match events per field. On the third no match event, the call is transferred to a CSA. The no match inputs assist in identifying callers that misunderstood the self service prompt and unique responses that the IVR application can use to improve field recognition coverage.

When the caller remains silent in response to an IVR application prompt, a no input event occurs. These events are represented by the no input parameters. Similar to the no match events per field, the application also accommodates a maximum of 3 no input events. No input events assist in identifying callers that were confused when prompted with the automated question. As a result, the caller remained silent. Instances when a caller response is longer than the allocated timeout period of the field are represented by the maximum speech timeout inputs. A maximum of 3 maximum speech timeout events per field are accommodated in the IVR application considered. These parameters indicate whether or not the timeout periods are adequate for callers to complete their responses. On the third no input and maximum speech timeout event, the call is also transferred to a CSA.

When a caller interrupts the IVR application prompt play, a barge-in event occurs. These instances are represented by the barge-in input. The caller disconnect and transfer to CSA inputs represent instances when a customer ends the call and when a call is transferred out of the IVR application to a CSA, respectively. A DTMF transfer input represents events when callers select the hash ("#") key on the phone. These inputs can also assist in determining the level of difficulty the caller experienced in the field.

The duration input indicates the total time the caller spent within a field. During the caller field interaction, the system error input parameter indicates whether or not a system error occurred. When the speech recognition probability is low at a field, in order to verify the recognized caller response is correct, a confirmation prompt is provided to the caller. The confirmed input represents the response of the caller to this prompt. This input also indicates whether or not a confirmation prompt had been presented to the caller.

In order to provide a summarized interpretation of caller behaviour at a field, the field performance, duration and recognition level output interaction classes are provided. Therefore, the field performance interaction class categorizes caller behaviour into good, acceptable, investigate or bad. The field duration as well as the recognition level classes illustrates 3 categories of performance, low, medium and high.

The field transfer and caller disconnect reason outputs indicate whether or not transfer to CSA and caller disconnect events occurred, respectively. Additionally, these interaction classes also indicate potential explanation for the transfer or caller disconnect event. Field difficultly attempt output determines the number of no match, no input and maximum speech timeout events that occurred during the interaction. Thus, these field classifier output classes provide further details of caller behaviour and assist in characterizing the caller experience. Experienced caller output classifies whether or not the caller is a regular user of the application and is therefore familiar with the application call flow. The experienced caller interaction class assists the call centre in determining the usage of the application.

A business intelligence solution that involved Extract, Transform and Load (ETL) processes is employed to create the datasets used in the development of the classifiers. The solution extracted and calculated information such as recognition confidence values, duration values as well as call completion information. Thereafter, the solution inserted the data into a database where specific rules were used to compute the required caller interaction information. Rules such as if no caller disconnect, transfer to CSA or system error occur, a maximum of 2 no inputs, no matches or maximum speech timeouts occur, the confidence level at the field is between 70% and 80%, the duration to complete the field is less than the average field duration and no field confirmation occurred, the field performance interaction class would be calculated as 'acceptable', are followed.

A 3 digit binary word notation is used to present the no match, no input and maximum speech timeout data to the field classifiers. For example, if no match 1, no match 2 and no match 3 events occur at a field, the binary notation will be '111'. In order to present the barge-in, caller disconnect, transfer to CSA, DTMF transfer and system error input information to the classifiers, a bit binary word notation is employed. A 2 digit binary word notation is used to present the confirmed input data to the classifier. The interpretation of the interaction output classes employed a similar binary notation scheme.

A normalization process is used to precondition the confidence and duration inputs. This process entails manipulating the datasets such that the values within the sets are between 0 and 1. The minimum and maximum values within the datasets are used to calculate the normalized values.

The data has been divided into training, validation and test sets to ensure that over-fitting and under-fitting were avoided. The training dataset is used to train the classifiers to identify the general classification groups within the data. The validation dataset is used to evaluate the classifier and the test data is utilized to confirm the classification capability of the developed models.

This research involved the development of 'Say account', 'Say amount', Select beneficiary' and 'Say confirmation' field classifiers. When a call enters the 'Say account' field element in an IVR pay beneficiary telephone banking application, the caller is prompted to say the account name or number. At the 'Say amount' field, the caller is requested to respond with the actual amount to pay the beneficiary. The 'Select beneficiary' field element is used to enable the caller to say the beneficiary name. The information recognized is read to the user and the caller is required to approve of the transaction within the 'Say confirmation' field. Due to the varying responses required at these fields, caller behaviour is unique per field. As a result, classifiers specifically trained on data relevant to the field are developed.

## 3. Fuzzy Inference System

Fuzzy Inference System (FIS) is a computational intelligent technique that computes outputs based on the fuzzy inference rules and present inputs. These techniques are based on fuzzy logic (Zadeh, 1994). FIS methods use fuzzification, inference and defuzzification processes. The mapping from the presented inputs to fuzzy sets defined in the corresponding universe is the process of fuzzification that results in fuzzy inputs. In order to generate the corresponding fuzzy outputs of these inputs, the decision making inference process employs fuzzy inference rules. The defuzzification process produces nonfuzzy outputs (Siler & Buckley, 2004).

Clustering of numerical data establishes the basis of many classification and system modelling applications. The objective of clustering is to locate natural groupings in a set of inputs with the intention of congregating similar inputs in the same region or class. When data clustering is used to compute fuzzy inference rules, the resultant rules are specifically tailored to the data. This is, therefore, an advantage when compared to FIS developed without clustering (Elwakdy et al., 2008).

In this research subtractive clustering is used to determine the optimal number and form of fuzzy inference rules. This clustering technique is a modified form of the Mountain Method for cluster estimation (Yen & Wang, 1999). Subtractive clustering considers each data point as a potential cluster centre and defines a measure of the potential of a data point (Chiu, 1994). The measure of potential for a certain data point is a function of its distances to all other data points. Therefore, a data point with many neighbouring points will have a high potential value. Once the potential of every data point has been computed, the point with the largest potential value is selected as the first cluster centre. Thereafter, in order to determine the next cluster and its centre, all data points in the vicinity of the first cluster centre that is determined by a radius of influence or cluster radius, is removed. This process is iteratively followed until all the input data are within a cluster radius of a cluster centre (Akbulut et al., 2004).

A subtractive clustering implementation is based on the cluster radius, squash factor, accept ratio and reject ratio parameters (Akbulut et al., 2004). Specifying a small cluster radius will result in many clusters within the data. Similarly, specifying a large cluster radius will yield few cluster centres. Additionally, in order to squash the potential for outlying points to be considered as part of a specific cluster, a squash factor is used. The squash factor is multiplied to cluster radius values. An accept ratio is used to set the potential above which another data point will be accepted as a cluster centre. This ratio is a fraction of the first cluster centre potential. A reject ratio is also used to specify the potential below which a data point will be rejected as a cluster centre. Similar to the acceptance ratio, the reject ratio is a fraction of the first cluster centre potential.

In this research, the subtractive clustering algorithm squash factor parameter is set to 1.25, thus indicating that only clusters that are far apart should be found. The accept radio parameter is set to 0.9, indicating that only data points that have a very strong potential for being a cluster centre are accepted. The reject radio parameter is set to 0.7, indicating that all data points without a strong potential are rejected.

Additionally, the research is concerned with a multi-dimensional caller behaviour classification problem. As a result, in order to determine the optimal FIS classifier, the cluster radius values for each input is optimized. GA, PSO and SA techniques are used to identify these optimal cluster radius values. The evaluation function of these methods is an error function that mapped the radius values to the accuracy values of the developed inference systems when presented with the validation and test datasets. The minimum value of these accuracies determined the fitness of the individual. Since this is a classification implementation, the accuracy of the FIS field classifiers developed can no longer be calculated utilizing the sum of square error of the difference between the target and the classifier output values. Instead, a confusion matrix is employed to identify the number of true and false classifications that are generated by the inference system developed. This is then utilized to calculate the true accuracy of the classifiers, using (1).

$$Accuracy = \sqrt{\frac{TP \times TN}{(TP + FN) \times (FP + TN)}} \qquad (1)$$

where *TP* is the true positive (1 classified as a 1); *TN* is the true negative (0 classified as a 0); *FN* is the false negative (1 classified as a 0) and *FP* is false positive (0 classified as a 1).
The sections to follow examine the GA, PSO and SA implementations.

## 4. Genetic Algorithm optimization

GA are known to be robust optimization procedures based on the mechanism of the natural evolution. After the original work conducted by Holland (Holland, 1975), GA were developed. These algorithms manage, maintain and manipulate populations of solutions utilizing a survival of the fittest strategy in their quest for an optimal solution. The populations of candidate solutions undergo a process of reproduction of individuals that advances individuals with better fitness values than the other individuals in the previous generation (Michalewicz, 1996). However, as evident in biological evolution, inferior individuals can also survive and reproduce.

The chromosome representation scheme defines the manner in which the problem is structured in the GA. An individual or chromosome comprises of a sequence of genes. Various types of chromosome representations such as binary digits, floating point numbers, integers, symbols and matrices exist.

In traditional GA, binary representation has been used for chromosomes that produce an even discrete depiction of the real optimization problem. Within binary coded GA, binary substrings representing each parameter with a desired precision are concatenated to form a chromosome. Therefore, a large number of variables in a real world problem would result in chromosomes encoded in long strings. However, natural representations are more efficient and may result in better solutions.

The floating point representation for chromosomes produces higher precision with more consistent results across replications (Michalewicz, 1996). In these real coded GA, a chromosome is coded as a finite length string of the real numbers corresponding to the real world problem variables. Real coded GA are robust, accurate as well as efficient because they are conceptually closest to the real world problem and moreover, the string length reduces to the number of variables. It has been reported that the real coded GA outperformed binary coded GA in many design problems (Janikow & Michalewicz, 1991). This research will determine if this is true in relation to the application problem of concern.

In order to produce successive generations, the selection of individuals is important within GA. The selection function identifies the individuals that will survive and proceed onto the next generation. A probabilistic selection is performed based upon the fitness of the individual such that the superior individuals have a higher possibility of being selected and therefore advancing onto the next generation. Several selection methods such as roulette wheel selection and its extensions, scaling techniques, tournament, elitist models and ranking selection techniques do exist (Houck et al., 1995). However, in this research normalized geometric ranking and tournament selection functions are employed within the GA implementations.

Ranking selection function methods utilize the evaluation function to map individual solutions to a completely ordered set, thus allowing minimization and negativity. When all individual solutions are sorted, these selection function techniques assign $P_i$ based on the rank of individual solution $i$. Normalized geometric ranking selection function, defines $P_i$ for each chromosome or individual solution using (2) (Houck et al., 1995).

$$P_i = q'(1-q)^{r-1} \tag{2}$$

$$q' = \frac{q}{1-(1-q)^P} \tag{3}$$

where $q$ is the probability of selecting the best individual; $r$ is the rank of the individual, where best equal 1; and $P$ is the population size.

The tournament selection function is a common selection mechanism used within GA (Booker et al., 1989). Similar to the ranking methods, the tournament selection function method only requires the evaluation function to map individual solutions to a partially ordered set. However, this selection technique does not assign probabilities to the individual solutions. This selection method begins by randomly selecting a number of individuals from the current population. The number of individuals is set by the tournament size. In this research, 3 individual solutions competed within each tournament. The selection method then compares the fitness of these individuals competing and inserts the individual solution with the best fitness value into the new population. This process is repeated until an appropriate population is achieved.

Crossover and mutation genetic operators are employed to provide basic search mechanisms for the GA. These operators produce new solutions based on existing individual solutions in the population. Crossover genetic operators use 2 individuals as parents to produce 2 new individual solutions. However, mutation operators alter 1 individual to yield a single new solution. Simple, arithmetic and heuristic crossover operators are usually utilized. Commonly employed mutation operators are boundary, non-uniform, uniform and multi-non-uniform operators. However, the types of operators used are dependent on the chromosome representation employed within the GA (Booker et al., 1989). As a result, binary coded GA employed binary mutation and simple cross over genetic operators. However, real coded GA utilized non-uniform mutation and arithmetic cross-over genetic operators (Houck et al., 1995). The binary mutation that entails flipping each bit in every individual within the population with probability $p_m$ is illustrated in (4) (Houck et al., 1995).

$$x'_i = \begin{cases} 1 - x_i & \text{, if } U(0,1) < p_m \\ x_i & \text{, otherwise} \end{cases} \quad (4)$$

(5) and (6) illustrates the simple crossover genetic operator that employs randomly generated number $r$ from a uniform distribution between 1 and $m$ to create 2 individuals (Houck et al., 1995).

$$x'_i = \begin{cases} x_i & \text{, if } i < r \\ y_i & \text{, otherwise} \end{cases} \quad (5)$$

$$y'_i = \begin{cases} y_i & \text{, if } i < r \\ x_i & \text{, otherwise} \end{cases} \quad (6)$$

As illustrated in (7), the non-uniform mutation genetic operator employed by the real coded GA selects 1 variable, $j$, and sets this variable equal to a non-uniform random number (Michalewicz, 1996).

$$x'_i = \begin{cases} x_i + (b_i - x_i)f(G) & \text{, if } r_1 < 0.5 \\ x_i - (x_i + a_i)f(G) & \text{, if } r_1 \geq 0.5 \\ x_i & \text{, otherwise} \end{cases} \quad (7)$$

where,
$$f(G) = (r_2(1 - \frac{G}{G_{max}}))^b \quad (8)$$

$a_i$ and $b_i$ are the lower and upper bound, respectively, for each variable $i$, $r_1$ and $r_2$ are uniform random numbers between 0 and 1; $G$ is the current generation, $G_{max}$ is the maximum number of generations and $b$ is the shape parameter.

(9) and (10) illustrates the arithmetic crossover operator that produces complimentary linear combinations of the parents, $\overline{X'}$ and $\overline{Y'}$ (Michalewicz, 1996).

$$\overline{X'} = r\overline{X} + (1-r)\overline{Y} \quad (9)$$
$$\overline{Y'} = (1-r)\overline{X} + r\overline{Y} \quad (10)$$

where, $\overline{X}$ and $\overline{Y}$ are two $m$-dimensional row vectors consisting of floating point valued individuals.

GA may begin with an initial population. This population can be randomly generated or derived from other methods. GA advance from generation to generation, ending when a termination criteria is achieved. The termination condition could be the maximum number of generations, population convergence criteria, lack of improvement of the best solution for a specified number of generations or achieving a specific targeted value within an objective function.

In this research, the GA implementations varied the number of individuals within the population. The GA produced 25 generations of these populations. Population sizes of 30, 100, 200, 300, 400 and 500 were considered. Table 2 illustrates the results of the most accurate binary coded and real coded GA implementations.

| FIS classifier | GA | Selection function | Population size | Accuracy | Evaluation function executions |
|---|---|---|---|---|---|
| 'Say account' | binary | normalized geometric ranking | 300 | 95.660% | 7,500 |
| | real | | 400 | 95.649% | 580 |
| | binary | Tournament | 30 | 95.660% | 750 |
| | real | | 300 | 95.649% | 490 |
| 'Say amount' | binary | normalized geometric ranking | 30 | 93.164% | 750 |
| | real | | 30 | 93.164% | 215 |
| | binary | Tournament | 30 | 93.164% | 750 |
| | real | | 30 | 93.164% | 222 |
| 'Select beneficiary' | binary | normalized geometric ranking | 100 | 95.777% | 2,500 |
| | real | | 100 | 95.680% | 280 |

|  | binary | Tournament | 100 | 95.747% | 2,500 |
|---|---|---|---|---|---|
|  | real |  | 100 | 95.680% | 290 |
| 'Say confirmation' | binary | normalized geometric ranking | 30 | 91.482% | 750 |
|  | real |  | 30 | 91.482% | 216 |
|  | binary | Tournament | 30 | 91.482% | 750 |
|  | real |  | 30 | 91.482% | 220 |

Table 2. Results of GA optimization implementation

When optimizing the FIS 'Say account' field classifier, the binary coded GA that employed geometric ranking selection function with a population size of 400 also converged to an optimal accuracy value of 95.660%. Similarly, the real coded GA that utilized the tournament selection function with a population size of 400 also converged to a solution value of 95.649%. However, in both implementations a population size of 300 executed the evaluation function less than a population size of 400. In regards to this FIS field classifier optimization, the binary coded GA that used the tournament selection function converged to an accuracy value of 95.660% with all population sizes considered. However, a population size of 30 called the evaluation function the least. It is also evident that the most accurate binary and real coded GA implementations converged to the same values of 95.660% and 95.649%, respectively.

In regards to FIS 'Say amount' classifier optimization, all GA implementations with all population sizes considered converged to an accuracy value of 93.164%. However, a population size of 30 executed the evaluation function the least. Similar results were obtained in the FIS 'Say confirmation' field classifier optimization.

When optimizing the FIS 'Select beneficiary' field classifier, all GA implementations with a population size of 100 yielded the most accurate results. As illustrated in Table 2, the most accurate real coded GA implementations converged to a solution value of 95.680%.

## 5. Particle Swarm optimization

PSO, created by Kennedy and Eberhart, is based on the co-operation of individuals, rather than the competitions among individuals (Kennedy & Eberhart, 1995). In the algorithm, a set of randomly generated solutions, the initial swarm, iteratively propagates through the search space towards an optimal solution. Members are known as particles and are not removed from the swarm. The swarm identifies the optimal solution using information determined about the search space and shared among all the particles. PSO is inspired by the capabilities of flocks of birds to adapt to their environment, to locate abundant food sources and avoid predators by employing an "information sharing" methodology. This technique provided an evolutionary advantage as it ensures that the swarm members access more information than that captured by their own senses.

When compared to global optimization algorithms such as SA, the major advantage of PSO is that it is resilient to local minima problems due to the large number of swarm members that comprise the particle swarm (Haiguo et. al, 2009; Jayabarathi et al., 2007; Ren et. al, 2008).

Several versions of the algorithm have been developed, incorporating techniques such as nearest-neighbour velocity matching and acceleration by distance (Eberhart et. al, 1996; Kennedy & Eberhart, 1995). However, after discovering that the algorithm can be used as a population-based optimizer, several parameters were removed using a trial and error process. The result was the first simple version of PSO (Eberhart et. al, 1996; Eberhart & Kennedy, 1995).

PSO explores the search space to identify the optimal solution by employing a population of particles that adapt by returning to previously successful regions (Eberhart & Shi, 2007). Particle movements are stochastic and are also influenced by experience of the particle and that of its neighbours. Each particle position represents a possible solution to the optimization problem. During iterations, the best solution achieved among all particles and the best solution achieved by each particle is recorded. These solutions are known as global best and personal best, respectively. At iterations, the velocity of each particle is also altered towards its personal best and the current global best, using (11). The particle positions are updated using (12). After several iterations, the swarm particles eventually cluster around regions with the fittest solutions.

$$v_i(t+1) = \omega v_i(t) + C_1 r(p_i - x_i(t)) + C_2 r(p_{bi} - x_i(t)) \tag{11}$$

$$x_i(t+1) = x_i(t) + v_i(t+1) \tag{12}$$

where $v_i$ is the current velocity of the $i^{th}$ particle, $p_i$ is personal best of $i^{th}$ particle, $p_{bi}$ is the global best, $\omega$ is the inertia weight, $C_1$ is the cognitive component, $C_2$ is the social component and $r$ represents uniformly distributed random values between 0 and 1. Similarly in (12), $x_i$ is the current position.

The random values ($r$) assists in maintaining diversity of the population, ensure good coverage of the search space and assist in avoiding local optima entrapment. The behaviour of PSO is influenced by the number of particles, inertia weight ($\omega$), cognitive component ($C_1$) and social component ($C_2$). A larger number of particles within the swarm increase social information exchange, thus increasing the probability of convergence to the global optimum.

The inertia weight is considered crucial in the convergence behaviour of the algorithm (Shi & Eberhart, 1998). The influence of the history of velocities on the current velocity is controlled by this parameter. The inertia weight provides a balance between global and local exploration capabilities of the swarm. A large inertia weight permits global exploration, thus allowing new regions within the search space to be examined. However, a small inertia weight assists local exploration, resulting in the fine-tuning of the current search region within the search space. As a result, an adequate inertia weight value can reduce the number of iterations required to locate the optimal solution. Research has discovered that good results can be achieved by initially setting the inertia weight value to a large value, and gradually reduce the value towards 0 (Shi & Eberhart, 1998). Therefore, this technique initially encourages global exploration of the search space and thereafter refining of the solution. In this research, this technique is used with an initial value of 0.9.

The appropriate selection of the cognitive and social component values can result in faster convergence to the optimal solution as well as prevent local minima convergences. Initial research proposed default value of 2 for both the cognitive component values (Kennedy, 1998). It has also been stated that a value of 0.5 for these components would produce better results (Parsopoulos & Vrahatis, 2002). However, it has also been suggested that a cognitive component value of 2.8 and a social component value of 1.3 maybe suitable settings (Carlisle & Dozier, 2001).

PSO is an evolutionary technique with a difference. Evolutionary techniques usually involve crossover, mutation and selection operators. This algorithm does not employee crossover operators. Information sharing in PSO occurs only among the own experience of the particle and the experience of the best particle within the swarm, rather than being inherited from fittest dependent individuals to descendants. Additionally, the directional position update method employed by PSO is similar to the GA.

PSO does not use the survival of the fitness strategy as it does not employ a direct selection function. As a result, during iterations, particles with lower fitness results can survive and potentially explore any point in the search space (Eberhart & Shi, 2007).

In this research, PSO employs the evaluation function described in the FIS section. The number of iterations is set to 25. PSO implementations with number of iterations set to 100, 200 and 300 were also considered. However, these implementations yielded similar results. PSO implementations with number of particles set to 30, 100, 200, 300, 400 and 500 were also considered. Additionally, for each of these implementations, the cognitive and social component values were varied. Both component values set to 0.5, 1, 1.5 and 2 were considered. The cognitive component value of 2.8 and social component value of 1.3 were also considered. Table 3 illustrates the results of the implementation.

| FIS classifier | $C_1$ | $C_2$ | Population size | Accuracy | Evaluation function executions |
|---|---|---|---|---|---|
| 'Say account' | 1 | 1 | 500 | 95.671% | 4,475 |
| 'Say amount' | 2 | 2 | 30 | 93.164% | 54 |
| 'Select beneficiary' | 1 | 1 | 300 | 95.767% | 2,561 |
| 'Say confirmation' | 2.8 | 1.3 | 100 | 91.482% | 151 |

Table 3. Results of PSO optimization implementation. $C_1$ and $C_2$ refer to the cognitive and social components, respectively.

As illustrated in Table3, the cognitive and social components values of 1 yielded the most optimal FIS 'Say account' and FIS 'Select beneficiary' field classifiers. All population sizes, cognitive and social components converged to the same optimal solution of 95.164% in the FIS 'Say amount' classifier optimization. However, the cognitive and social component values of 2 executed the evaluation function the least. When optimizing the FIS 'Say confirmation' classifier, majority of the PSO implementations converged to a solution value of 91.482%. However, the PSO implementation with population size, cognitive and social components set to 100, 2.8 and 1.3, respectively, executed the evaluation function the least.

## 6. Simulated Annealing optimization

Annealing is the process of heating up a solid metal and thereafter cooling it slowly until crystallization occurs. At very high temperatures, atoms within the metal have high levels of energy, thus providing the atoms freedom to restructure. As the temperature is reduced, the levels of energy decrease, until a state of minimum energy is achieved. The resultant frozen metal crystallization characteristic is determined by two annealing parameters; the initial temperature at which the process begins and the rate at which the temperature is reduced. The rate at which the temperature is reduced should be low, to allow the atoms in the metal to settle and provide sufficient time to form a crystal lattice with minimum internal energy. However, a slow rate would lead to long solidifying process duration. In order to reduce this process duration, a low initial temperature can be considered. The initial temperature should be sufficient to allow the atoms freedom to rearrange their positions.

SA emulates this process (Laarhoven & Aarts, 1987). The algorithm employs a high initial temperature value, allowing the inputs to assume a wide range of values. At iterations, the temperature is gradually reduced, restricting the range of input values. This process often results in the algorithm producing an optimal solution, similar to the metal achieving an optimal crystal structure through the annealing process. Due to the ability of SA accepting not only the better solutions but also inferior solutions with a given probability, the algorithm avoids local minima.

It has been proven that the algorithm converges to the global optima. The convergence proof employs a very slow cooling rate that sets the initial condition to a sufficiently large temperature. Thereafter, the temperature is reduced by $T_k = T_0/\log(k)$, where $k$ is bound by the number of iterations (Ingber & Rosen, 1992). The proof illustrates a longer cooling rate results in better quality solutions. However, a very slow cooling rate is impractical.

SA comprises of three functional relationships; a probability density $g(x,T)$, an acceptance function $h(\Delta E, T)$ and the annealing schedule function $T(k)$ with the time step $k$ (Ingber, 1996). Initially, the algorithm begins at a randomly chosen point of which the fitness value is computed using an evaluation function. Thereafter, a new point is selected from a random number generator with a probability density $g(x)$. If the fitness value of this point is better than the previously selected point, the new point is chosen. However, if the fitness value is worse than the previously selected point, the new point is accepted by a probability $h(\Delta E, T)$. A point is always selected based on the best point. As the number of iterations increase, the probabilities for large deviations from the best point and for acceptance decrease. Therefore, the algorithm encourages the exploration of distant points at the beginning, and does not generate or accept distant points as the temperature is reduced.

In this research, standard Boltzmann annealing is used (Ingber, 1996). (13), (14) and (15) illustrate the probability density, acceptance function and annealing schedule function, respectively.

$$g(x,T) = (2\pi T)^{-\frac{D}{2}} \exp\left(-\frac{(\Delta x)^2}{2T}\right) \qquad (13)$$

$$h(\Delta E, T) = \frac{1}{1 + \exp\left(\frac{\Delta E}{T}\right)} \qquad (14)$$

$$T(k) = \frac{T_0}{\ln k} \qquad (15)$$

where, $D$ is the dimension of the search space; $T$ is the temperature; $T_0$ is the initial temperature; $\Delta x$ is the difference between point $x_{new}$ and $x$; $\Delta E$ is the energy difference between the fitness function value of point $x_{new}$ and fitness function value of point $x$.

Additionally, in this research exponential and fast annealing schedules were also considered, as defined in (16) and (17), respectively.

$$T(k) = T_0 \times 0.95^k \tag{16}$$

$$T(k) = \frac{T_0}{k} \tag{17}$$

In order to determine the optimal SA implementation, the number of iterations is also varied from 1 to 100. The initial temperature values of 1 and 100 were also considered. Table 4 illustrates the results of the SA implementation.

| FIS classifier | Iterations | $T(k)$ | $T_0$ | Accuracy | Evaluation function executions |
|---|---|---|---|---|---|
| 'Say account' | 86 | Boltzmann | 1 | 95.659% | 88 |
| | *78* | *Exponential* | *1* | *95.660%* | *80* |
| | 99 | Fast | 1 | 95.659% | 101 |
| 'Say amount' | 1 | Boltzmann | All | 93.164% | 3 |
| | 1 | Exponential | All | 93.164% | 3 |
| | 1 | Fast | All | 93.164% | 3 |
| 'Select beneficiary' | *100* | *Boltzmann* | *1* | *95.699%* | *102* |
| | 9 | Exponential | 1 | 95.680% | 11 |
| | 67 | Fast | 1 | 95.684% | 69 |
| 'Say confirmation' | *4* | *Boltzmann* | *1* | *91.482%* | *6* |
| | *4* | *Exponential* | *1* | *91.482%* | *6* |
| | 5 | Fast | 1 | 91.482% | 7 |

Table 4. Results of SA optimization implementation. $T(k)$ and $T_0$ refer to the annealing schedule and initial temperature values, respectively.

As illustrated in Table 4, the initial temperature value of 1 achieved the best results for FIS 'Say account' and 'Select beneficiary' field classifier optimizations. However, the initial temperature value of 100 did converge to an optimal solution at lower iterations in both FIS field classifier optimizations. The exponential and Boltzmann annealing schedules with an initial temperature of 1 yielded the most accurate FIS 'Say account' and 'Select beneficiary' field classifiers.

When optimizing the FIS 'Say amount' and 'Say confirmation' field classifiers, both initial temperature values considered yielded the same optimal accuracy value of 93.164% and 91.482%, respectively. Additionally, varying the number of iterations for FIS 'Say amount' optimization did not improve the accuracy value. All the annealing schedules considered in this optimization also yielded the same optimal value. In regards to FIS 'Say confirmation' optimization, the initial temperature value of 1 achieved the optimal solution at lower iterations. The Boltzmann and exponential annealing schedules yielded the optimal solutions using only 6 iterations.

In majority of the SA implementations considered, the initial temperature of 100 did not yield the most accurate FIS field classifier. As a result, it can be concluded that this initial temperature value result in the exploration of points very far from the optimal value and, therefore, does not converge to the optimal value as this temperature value is reduced.

## 7. Comparison of optimization results

The optimization algorithms considered in this research have yielded FIS field classifiers with accuracies above 91%. Table 5 illustrates the most accurate field classifiers produced by the optimization techniques considered. PSO produced the most accurate FIS 'Say account' field classifier. All the optimization algorithms considered resulted in FIS 'Say amount' and 'Say confirmation' field classifiers with accuracy values of 93.164% and 91.482%, respectively. The binary coded GA that used normalized geometric ranking selection function yielded the most accurate FIS 'Select beneficiary' field classifier. However, PSO produced a FIS 'Select beneficiary' field classifier that is only 0.01% less accurate.

| FIS classifier | Algorithm | Accuracy | Evaluation function executions |
|---|---|---|---|
| 'Say account' | GA | 95.660% | 750 |
|  | **PSO** | **95.671%** | **4,475** |
|  | SA | 95.660% | 80 |
| 'Say amount' | GA | 93.164% | 215 |
|  | PSO | 93.164% | 54 |
|  | **SA** | **93.164%** | **3** |
| 'Select beneficiary' | **GA** | **95.777%** | **2,500** |
|  | PSO | 95.767% | 2,561 |
|  | SA | 95.699% | 102 |
| 'Say confirmation' | GA | 91.482% | 216 |
|  | PSO | 91.482% | 151 |
|  | **SA** | **91.482%** | **6** |

Table 5. Comparison of optimization algorithms considered

In this research, GA, PSO and SA are compared in terms of computational efficiency and quality of solution. Computational efficiency, in this context, is defined as the number of times the evaluation function is executed. Quality of solution is the confirmation that the optimal FIS field classifier produced is truly the most optimal solution.

In regards to computational efficiency, the most accurate SA converged to an optimal solution using the least number of evaluation function executions. This is true for all the FIS field classifier optimizations. As illustrated in Table 5, when comparing the most accurate PSO, the most accurate GA converged to an optimal solution using less number of evaluation function executions in the FIS 'Say account' and 'Select beneficiary' field classifier optimizations.

In order to confirm the quality of solution, sensitivity and specificity of the most accurate field classifiers produced by the algorithms considered is calculated using (18) and (19), respectively.

$$Sensitivity = \frac{TP}{TP + FN} \qquad (18)$$

$$Specificity = \frac{TN}{TN + FP} \qquad (19)$$

Sensitivity is defined as the probability that the FIS classifier categorizes a set of caller behaviour inputs to the correct specific output interaction classes. This performance measure describes the effectiveness of the FIS classifier at classifying interaction classes correctly. Specificity is defined as the probability that the field classifier indicates that a set of inputs does not correctly belong to specific output interaction classes. This measure characterizes the performance of the field classifier at discarding interaction classes that are not relevant to the presented inputs. Table 6 illustrates the performance measures used in evaluating the field classifiers produced by the optimization techniques.

| FIS classifier | Algorithm | Accuracy (%) | Sensitivity (%) | Specificity (%) |
|---|---|---|---|---|
| | | **Validation dataset** | | |
| 'Say account' | GA | 95.660 | 94.452 | 96.884 |
| | *PSO* | *95.671* | *94.474* | *96.884* |
| | SA | 95.660 | 94.452 | 96.884 |
| 'Say amount' | GA | 93.962 | 91.561 | 96.425 |
| | PSO | 93.962 | 91.561 | 96.425 |
| | *SA* | *93.962* | *91.561* | *96.425* |
| 'Select beneficiary' | *GA* | *95.777* | *93.333* | *98.285* |
| | PSO | 95.767 | 93.264 | 98.336 |
| | SA | 95.699 | 93.109 | 98.361 |
| 'Say confirmation' | *GA* | *91.482* | *85.815* | *97.524* |
| | PSO | 91.482 | 85.815 | 97.524 |
| | SA | 91.482 | 85.815 | 97.524 |
| | | **Test dataset** | | |
| 'Say account' | GA | 96.641 | 96.062 | 97.224 |
| | *PSO* | *96.641* | *96.062* | *97.224* |
| | SA | 96.624 | 96.038 | 97.213 |
| 'Say amount' | GA | 93.164 | 89.800 | 96.654 |
| | PSO | 93.164 | 89.800 | 96.654 |
| | *SA* | *93.164* | *89.800* | *96.654* |
| 'Select beneficiary' | *GA* | *95.951* | *93.963* | *97.980* |
| | PSO | 95.965 | 93.702 | 98.284 |
| | SA | 96.015 | 93.654 | 98.437 |
| 'Say confirmation' | *GA* | *92.147* | *86.901* | *97.711* |
| | PSO | 92.136 | 86.879 | 97.711 |
| | SA | 92.136 | 86.879 | 97.711 |

Table 6. Performance measures of optimization algorithms considered

When presented with the test dataset, the FIS 'Say account' field classifiers produced by PSO and GA yielded the same accuracy values. However, the field classifier optimized by PSO yielded better results with the validation dataset inputs and is therefore the preferred method in regards to the FIS 'Say account' field classifiers as well as quality of solution.

It is evident from Table 6 that the FIS 'Say amount' field classifiers optimized by all the algorithms considered produced the same results for both validation and test datasets. As a result, all the algorithms perform the same in regards to quality of solution in FIS 'Say amount' field classifier optimization.

The FIS 'Select beneficiary' field classifier optimized by the GA performed the best on validation dataset. However, PSO produced a FIS 'Select beneficiary' field classifier that is 0.014% more accurate on test dataset and is therefore the preferred method in relation to quality of solution.

When presented with the validation dataset, the FIS 'Say confirmation' field classifiers optimized by all the algorithms considered yielded the same results. However, the field classifier produced by the GA yielded the best results on test 'Say confirmation' data and is therefore the preferred technique in regards to quality of solution.

All the field classifiers produced in this research have high positive and negative classification rates. This is evident from the high sensitivity and specificity performance metric values. However, the number of evaluation function executions varies significantly.

As illustrated in Table 5, SA executes the evaluation function significantly less than PSO and GA. Additionally, the FIS field classifiers produced by SA are only on average 0.022% less accurate than PSO and GA optimized classifiers. Therefore, if computational efficiency is a priority, SA would be preferred for all datasets considered in this research and would also result in acceptable accuracy.

However, if small accuracy improvements are important in the system, PSO is the preferred optimization method for FIS 'Say account' and 'Select beneficiary' field classifiers. Similarly, GA is the preferred technique for FIS 'Say confirmation' field classifier optimization.

Research conducted has concluded that PSO is more effective than GA (Panduro et. al, 2009) and vice versa (Jones, 2006). Similarly, research has been conducted that has stated that SA is more effective than GA (Lahitinen et. al, 1996) and vice versa (Manikas & Cain, 1996). This also applies to PSO (Ethni et. al, 2009). As a result, the performance of these techniques is very dependent on the problem under consideration.

Table 7 illustrates the results of previous FIS field classifiers developed (Patel & Marwala, 2010). Subtractive clustering was also used to generate the FIS inference rules. However, each input or dimension employed the same cluster radius.

| Classifier | FIS | Accuracy (%) | Sensitivity (%) | Specificity (%) |
|---|---|---|---|---|
| Validation dataset | | | | |
| 'Say account' | Previous | 80.680 | 72.000 | 99.430 |
| | *Optimized* | *95.671* | *94.474* | *96.884* |
| 'Say amount' | Previous | 82.650 | 70.220 | 98.600 |
| | *Optimized* | *93.962* | *91.561* | *96.425* |
| 'Select beneficiary' | Previous | 78.430 | 61.170 | 99.000 |
| | *Optimized* | *95.777* | *93.333* | *98.285* |
| 'Say confirmation' | Previous | 79.510 | 71.380 | 99.580 |
| | *Optimized* | *91.482* | *85.815* | *97.524* |
| Test dataset | | | | |
| 'Say account' | Previous | 80.770 | 66.150 | 98.630 |
| | *Optimized* | *96.641* | *96.062* | *97.224* |
| | Previous | 82.540 | 68.470 | 99.510 |
| | *Optimized* | *93.164* | *89.800* | *96.654* |
| 'Select | Previous | 77.820 | 62.730 | 99.080 |
| beneficiary' | *Optimized* | *95.951* | *93.963* | *97.980* |
| | Previous | 79.470 | 65.760 | 96.040 |
| | *Optimized* | *92.147* | *86.901* | *97.711* |

Table 7. Comparison of previously developed and current FIS classifier results

It is evident from Table 7 that the optimization techniques considered in this research has made a significant improvement on the accuracy of the FIS field classifiers. On average, there is an increase in accuracy of 13.906% and 14.326% on the validation and test datasets, respectively.

## 8. Conclusion

The optimization techniques considered in this research produced highly accurate FIS 'Say account', 'Say amount', 'Select beneficiary' and 'Say confirmation' field classifiers that were optimized using GA, PSO and SA methods. Subtractive clustering is used to generate the FIS inference rules, thus yielding resultant rules specifically tailored to the data. An error function that mapped cluster radii to the FIS field classifier accuracy values of validation and test datasets formed the evaluation function employed within these optimization methods.

Binary and real coded GA that employed normalized geometric and tournament selection functions were used to compute optimal cluster radius values. These GA implementations also entailed varying the population size. When optimizing the FIS 'Say account' and 'Select beneficiary' field classifiers, the binary coded GA outperformed the real coded GA. Both GA encodings produced the same FIS 'Say amount' and 'Say confirmation' classifier results. However, the real coded GA executed the evaluation function the least.

PSO implementations used various cognitive and social component values. Additionally, the PSO implementation process also involved varying the population size. Cognitive and social components values of 1 produced the most optimal FIS 'Say account' and 'Select beneficiary' field classifiers. When optimizing the FIS 'Say amount' field classifier, the cognitive and social component values of 2 executed the evaluation function the least. When optimizing PSO FIS 'Say confirmation' classifier, a population size, cognitive and social components set to 100, 2.8 and 1.3, respectively, executed the evaluation function the least.

In this research, Boltzmann, exponential and fast annealing schedules were considered in the SA implementation. The number of iterations is also varied. Initial temperature values of 1 and 100 were also considered. In majority of the FIS classifier optimizations, an initial temperature value of 1 achieved better results than an initial temperature value of 100. An exponential and Boltzmann annealing schedule achieved good results for FIS 'Say account' and 'Select beneficiary' classifier optimizations, respectively. Both exponential and Boltzmann annealing schedules yielded the same results in FIS 'Say confirmation' classifier optimization. All annealing schedules and initial temperature values achieved the same results in 'Say amount' field optimization.

In this research, GA, PSO and SA methods are compared in regards to computational efficiency and quality of solution. If computational efficiency is a priority, the SA technique is preferred for all FIS field classifiers yielding acceptable accuracy values. However, if small accuracy improvements are more important than computational efficiency, PSO method is preferred for FIS 'Say account' and 'Select beneficiary' field classifier optimizations. GA is the preferred technique for FIS 'Say confirmation' field classifier optimization. All the optimization methods considered yielded the same accuracy values for the FIS 'Say amount' field classifier. As a result, due to the computational efficiency of the SA technique, it is the preferred method for FIS 'Say amount' classifier optimization.